\PassOptionsToPackage{usenames,dvipsnames,table}{xcolor}
\documentclass[11pt,a4paper]{article}
\usepackage{emnlp-ijcnlp-2019}
\usepackage{times}
\usepackage{latexsym}
\usepackage{url}
\usepackage{rst}
\usepackage{amssymb}
\usepackage{times}
\usepackage{latexsym}
\usepackage{amsmath}
\usepackage{multirow}
\usepackage{url}
\usepackage{amsfonts}
\usepackage{booktabs}
\usepackage{rotating}
\usepackage{subfig}
\usepackage{array}
\usepackage{booktabs}
\usepackage{amsfonts}
\usepackage{enumitem}% http://ctan.org/pkg/enumitem
\usepackage{makecell}
\usepackage{xcolor}
\usepackage{caption}
\usepackage{array}
\usepackage[table]{xcolor}% http://ctan.org/pkg/xcolor
\usepackage{setspace}
\usepackage[toc,page]{appendix}
\usepackage{multirow}
\usepackage{tabularx}
\usepackage{xspace} 
\usepackage{tipa}
\definecolor{vlgray}{gray}{0.6}
\usepackage{soul}

\newcolumntype{x}[1]{>{\centering\arraybackslash\hspace{0pt}}p{#1}}
\usepackage{floatrow}
\newfloatcommand{capbtabbox}{table}[][\FBwidth]

\usepackage{blindtext}

\usepackage{etoolbox}
\newtoggle{draft}
\toggletrue{draft}
\iftoggle{draft}{
\newcommand{\dk}[1]{\textcolor{Maroon}{[#1 \textsc{--dk}]}}
\newcommand{\ed}[1]{\textcolor{Blue}{[#1 \textsc{--ed}]}}
\newcommand{\hh}[1]{\textcolor{cyan}{[#1 \textsc{--hh}]}}
\newcommand{\tj}[1]{\textcolor{BurntOrange}{[#1 \textsc{--tj}]}}
}{
\newcommand{\dk}[1]{}
\newcommand{\ed}[1]{}
\newcommand{\hh}[1]{}
\newcommand{\tj}[1]{}
}

\def\BState{\State\hskip-\ALG@thistlm}

\usepackage{etoolbox}
\preto\align{\par\nobreak\normalsize\noindent}
\expandafter\preto\csname align*\endcsname{\par\nobreak\normalsize\noindent}

\newcommand{\method}{FlowNet\xspace}

\newcolumntype{P}[1]{>{\centering\arraybackslash}p{#1}}

\aclfinalcopy % Uncomment this line for the final submission

\title{
Linguistic Versus Latent Relations \\for Modeling Coherent Flow in Paragraphs
}

\author{
Dongyeop Kang \quad Hiroaki Hayashi \quad Alan W Black \quad Eduard Hovy\\
Carnegie Mellon University, Pittsburgh, PA, USA \\
{\tt $\{$dongyeok,hiroakih,awb,hovy$\}$@cs.cmu.edu   }
}

\date{}

\begin{document}
\maketitle

\begin{abstract}
% \hh{I think we need to revise the abstract later, by making clear
% that a) we focus on document-level LM b) we are interested in cross-sentencial information propagation c) we examine learnable latent version and linguistically labeled version.}
% \dk{Agree with (b) and (c) but not (a). If (a), define "document" as a multi-sentences such as a paragraph}
Generating a long, coherent text such as a paragraph requires a high-level control of different levels of relations between sentences (e.g., tense, coreference).
%In general, we call sentences put together in a logical order
We call such a logical connection between sentences as a (paragraph) \textit{flow}.
In order to produce a coherent flow of text, we explore two forms of inter-sentential relations in a paragraph:
one is a human-created linguistical relation that forms a structure (e.g., discourse tree) and the other is a relation from latent representation learned from the sentences themselves.
%which we compare them in generation tasks
Our two proposed models incorporate each form of relations into document-level language models: the former is a supervised model that jointly learns a language model as well as discourse relation prediction, and the latter is an unsupervised model that is hierarchically conditioned by a recurrent neural network (RNN) over the latent information.
%linguistic discourse relations (\explicit) and the latent delta relations (\implicit), and compare them in various generation tasks.\hh{suddenly delta appeared}
% Each form has its own benefit:
% while linguistic relations help narrow down the search space of the language model with explicitly labeled relations, the latent relations implicitly capture topic-invariant flow (e.g., general writing scheme) over sentences.
% Since typical discourse relations do not always ensure lyrical flow, we \dk{TODO}.
%by a simple subtraction operation of two sentence representations.
% Both linguistic and latent relations help narrow down the search space of the language model and model the underlying consistent flow in the text in either explicitly with a discourse relation from or implicitly with a delta representation. \hh{what is search space of a language model?}
Our proposed models with both forms of relations outperform the baselines in partially conditioned paragraph generation task.
Our codes and data are publicly available\footnote{\url{https://github.com/dykang/flownet}}.
% \dk{Daniel's comments: add unsupervised / supervised in modeling side}
\end{abstract}

%%%%%%%%%%%%%%%%%%%%%%%%%%%%%%%%%%%%%%%%%%%%%%%%%%%%%%
\section{Introduction}\label{sec:intro}
%%%%%%%%%%%%%%%%%%%%%%%%%%%%%%%%%%%%%%%%%%%%%%%%%%%%%%
% \dk{
% bring word level subtration at tje beginning to remove the topics
% topic transition. good indicator: positiveness.
% }

When composing multiple sentences into a paragraph, as in novels or academic papers, we often make design decisions in advance~\cite{byrne1979teaching} such as topic introduction and content ordering to ensure better coherence of the text.
% in order to write a coherent and persuasive text.
% such as what to say about, what order of sentences to place, how to control level of abstraction and more.
For instance, \citet{mckeown1985discourse,gopen1990science} proposed effective patterns for scientific writing:
%a topical sentence at first followed by some example sentences and a concluding sentence with stronger stance at the end.
a hypothesis at first, followed by supporting sentences to validate the hypothesis, and lastly a concluding sentence.
We call such a logical connection between sentences in a written paragraph as a \textit{flow}.
%communication, goals,
A coherent flow between sentences requires an understanding of various factors including tense, coreference, plans \cite{appelt1982planning,hovy1991approaches}, scripts \cite{tomkins1978script} and several others.
We focus on the paragraph-level plan between sentences.

% \begin{figure}[t]
% \centering\vspace{0mm}
% \includegraphics[trim={0cm 75mm 175mm 0cm},clip,width=0.75\linewidth]{figs/{motivation}.pdf}\vspace{0mm}
% \caption{
% % \hh{URGENT: For consistency, please use $\Delta$ rather than $d$.}
% Modeling an inter-sentential flow using a linguistically labeled information (\explicit) (e.g., a discourse relation) or a learnable latent information (\implicit). We are curious which forms of relations are more useful for coherent text generation.
% %modeled by a simple operation such as subtraction of two adjacent sentences.
% } \label{fig:motivation}\vspace{-5mm}
% \end{figure}

%\hh{IMPORTANT: deal with ``plan'' mentions. ``type of relations'' should change to ``form of relations''. (``Type'' fits better with each of discourse relation type, e.g. cause/elaboration.)}\dk{Good point. Changed all of them in text accordingly.
% This example sets a guideline on typical content coherence construction.
% Likewise, coherence can be enforced in various ways and granularities, ranging from relatively simpler factors like tense or coreference, to semantically more complex plans \cite{hovy1991approaches}, scripts \cite{tomkins1978script} and others.
In text planning, underlying relations in text are broadly categorized into two forms:
an explicit human-defined relation (e.g., a discourse tree) \cite{reiter2000building} or an implicitly learned latent relation \cite{yang2016hierarchical}.
While the former is defined and manuallly annotated based on linguistic theories, the latter is simply determinable from how people in fact put sentences together. % % \hh{Do you have any ref about plan, defined as relations between sentences?}\dk{AFAIK, I can't find any ref. Dont' need a specific citation for this because it's just scope of our work.}
In this work, we provide an empirical comparison between a linguistically-informed and a latent form of relations in context of a paragraph generation.

We compare the effectiveness of the two forms of relations using language modeling for paragraph generation.
Due to the different characteristics of the two forms, we employ comparable but different components in addition to the base language model.
For linguistic relations (e.g., discourse), we cast the problem
into multi-task learning of supervised language modeling and discourse relation prediction. % using conditional random field (CRF) \cite{Lafferty2001ConditionalRF}.
On the other hand, for latent relations, we learn an unsupervised hierarchical language model that is hierarchically conditioned by RNNs over linear operations between sentences.

We evaluate our models on partial paragraph generation task; producing the rest of text in a paragraph given some context of text.
We observe that linguistically annotated discourse relations help produce more coherent text than the latent relations, followed by other baselines.
% In the partial paragraph generation task we defined; given some context of text, producing the rest of text in a paragraph, we observe that linguistically annotated discourse relations help produce more coherent text than the latent relations, followed by other baselines.
% The hierarchical modeling of the delta relations helps avoid the repetition issue of the prior unsupervised language models.

% To assess the models' ability of capturing coherence, we propose two tasks: partial paragraph generation and sentence ordering.
% Our model with either linguistic or latent relation outperforms the baselines for the generation and the ordering task.

% We first collect large corpora of paragraphs from news articles, novel books and academic papers.

% \input{02analysis.tex}

%%%%%%%%%%%%%%%%%%%%%%%%%%%%%%%%%%%%%%%%%%%%%%%%%%%%%%
\section{Related Work}\label{sec:related}
%%%%%%%%%%%%%%%%%%%%%%%%%%%%%%%%%%%%%%%%%%%%%%%%%%%%%%
There has been a variety of NLG systems that incorporate additional information between sentences~\cite{appelt1982planning,reiter2000building,gatt2018survey} which can be broadly categorized into two forms: linguistic and latent.

\textbf{Linguistic} relations are explicitly represented as external labels in the form of predefined rules or plans, formats, knowledge base, discourse parses, and more.
\citet{Hovy1985IntegratingTP,hovy1990pragmatics,dalianis1996aggregation} integrated text planning in generation, where the plans are considered in knowledge, formatted rules and so forth. 
However, they are limited to small scale (i.e. few examples) and hand-written rules.
%,kang2017detecting,kang18acl
\citet{kang2017detecting,gardent2017creating,kang18acl,Wang2018DescribingAK} used an external knowledge base to micro-planning for generating a corresponding text, while our work focuses on comparing two forms of relations from the text itself.
% \citet{kang2017detecting,kang18acl} proposed to generate explanations using causality relations or entailment constraints between sentences.
% Our work proposes more generalized models that can employ different kinds of relations on text generation. 
%\hh{You say, these work are based on labeled relations. and they are bad because they are based on labels?, also, Gardent's work can scale to larger dataset. He did crowdsourcing.} \dk{Fixed}
%\hh{You didn't say that the previous work couldn't incorporate different kinds of relations, so your work isn't contrasted much.}\dk{divide the text into two pieces and add one to the first paragraph and other to here.}
% type of explicit relations are 
% One of our precious resource for modeling consistency is discourse relations.
%\hh{No. it's for modeling document composition.}\dk{You're right. See below}

\citet{moore1993planning,young1994dpocl} utilized discourse structures such as rhetorical structure theory (RST)~\cite{mann1988rhetorical} for parsing a document. 
%\hh{Oh how? isn't this pretty similar to what discourse flow is doing?}\dk{Fixed.}
A script~\cite{tomkins1978script} is another structured representation that describes a typical sequence of events in a particular context.
\citet{zhang2016variational,ji2014representation} proposed better discourse parsers using neural networks.
The prior works, however, used the discourse representations to describe the structure of the paragraph, while we focus on applicability of the discourse relations to language generation.

%,li2014recursive,feng2014linear
\textbf{Latent} relations use implicit information in a document such as hierarchical structure of the document:
\citet{lin2015hierarchical,chung2016hierarchical} used hierarchical RNN for modeling a document. 
Similarly, the hierarchical model can be extended to other variants such as attention~\cite{yang2016hierarchical}, encoder-decoder framework~\cite{serban2017hierarchical,sordoni2015hierarchical}, auto-encoding~\cite{li2015hierarchical}, and multiscale~\cite{chung2016hierarchical}. 
However, the hierarchical recurrence of sentences, which is dependent on topics, are less likely modeling a flow of a document.

We further summarize the fundamental differences between the two forms of relations in Appendix.

\section{\method: Language Modeling with Inter-sentential Relations}\label{sec:flow}
%%%%%%%%%%%%%%%%%%%%%%%%%%%%%%%%%%%%%%%%%%%%%%%%%%%%%%%%%%%%%%%%%%%%

\begin{figure*}[t]
\centering\vspace{-2mm}
{
\subfloat[][Discourse-driven]{
\includegraphics[trim={0mm 109mm 177mm 0mm},clip,width=0.43\linewidth]{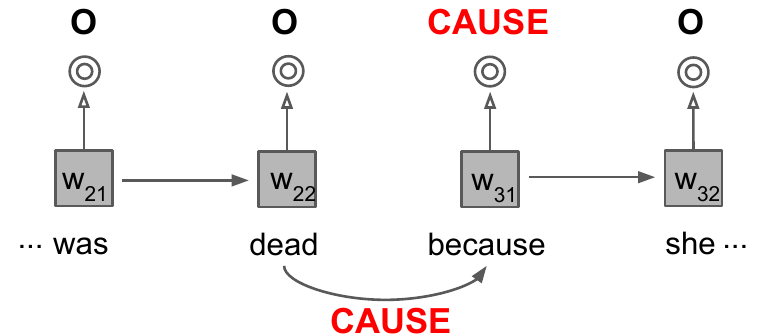}
}
\quad\quad\quad
\subfloat[][Delta-driven]{\includegraphics[trim={0mm 168mm 185mm 0mm},clip,width=0.43\linewidth]{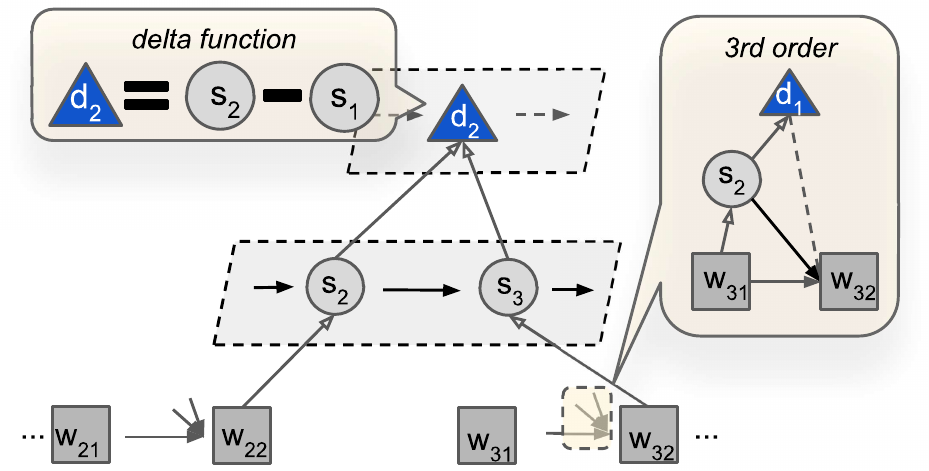}}
}\vspace{0mm}
\caption{\method with linguistic (i.e., discourse) versus latent (i.e., delta) relation. (a) For each word, a form of discourse relation and next word are jointly predicted using CRF ($\circledcirc$) and language model, respectively. (b) Decoding $w_{i}$ is conditioned on previous word ($w_{i-1}$), previous sentence ($s_{i-1}$), and delta between two previous sentences ($d_{i-2}$). Best viewed in color.
% \tj{on (b), d2 = s2 - s1 part should be changed to d2 = s3 - s2 following the definition of d}
\vspace{0mm}
} \label{fig:framework}
\vspace{0mm}
\end{figure*}

We propose language models that incorporate each relation to capture a high-level flow of text. %\hh{I guess you need to define flow first...}\dk{good point. See below.}
% In contrast to a sequence of sentences, a flow is modeling topic-invariant sequence on top of them, conditioned on either explicit discourse relations or implicit delta representations.

%%%%%%%%%%%%%%%%%%%%
\subsection{Discourse-driven \method}\label{sec:lmdiscourse}
%%%%%%%%%%%%%%%%%%%%
As a linguistic relation, we employ RST~\cite{mann1988rhetorical} trees to represent discourse connections in the text.
For simplicity, we limit usage of the discourse trees by only considering relations between adjacent phrases\footnote{The full discourse tree can be incorporated using other types of language model such as \citet{tai2015improved}.}: relations are inserted between adjacent phrases and represented as a flattened sequence of phrases and relations.
If two consecutive RST relations are given, the deeper level of relation is chosen. If the central elementary discourse unit (EDU) or phrase is after its dependent, the relation is excluded.
We consider each sequence of the flattened discourse relations as a writing flow.
For example, people often write a text by elaborating basic information (\texttt{Elaboration}) and then describing a following statement attributed to the information (\texttt{Attribution}).
% If we extend it to a full sequence of RST relations in a paragraph and incorporate them into text generation, we may model a partial high-level flow of text conditioned by the RST relations.

We view discourse relations as additional labels to predict at the same time we predict next words in language modeling.
Specifically, we propose to jointly train a model that predicts a sequence of words and a sequence of RST labels by taking advantage of shared representations, following previous sequence labeling problems such as named entity recognition~\cite{collobert2011natural} and part-of-speech tagging~\cite{huang2015bidirectional}.
Note that the RST relations are only used during training to obtain better representation for the two tasks, but not at test time.
% At testing time, the model takes advantages from the shared representation.

Figure~\ref{fig:framework}(a) shows our \method using discourse relations. Let a paragraph be a sequence of sentences $D$=$\{s_1, s_2, \ldots, s_M\}$.
% The model encodes each phrase $c = g(s_{i-1})$ in a RST parse where $g$ is a BiLSTM encoder such as $h_i = [ \overrightarrow{h_i} ; \overleftarrow{h_i} ]$.
This model treats adjacent sentences as pairs for learning the standard seq2seq model.
% The model encodes words in the input sentence $s_{i-1}$ to obtain context $c = \text{RNN}(s_{i-1})$, and feeds it to the decoder $h_t = f(w_{i,t-1}, h_{t-1}, c)$
% to generate the next sentence $s_i=\{w_{i1},\ldots w_{iT_i}\}$ where $g$ and $f$ are RNNs, and $T_i$ represents the sentence length.
% %In unconditional language modeling, the model directly decodes from the input text, while in conditioned text generation (i.e., generate a text given a context) we encode the context %$c$ \hh{isn't this c?}\dk{Right. Corrected.}
%first and then decode it.
%We evaluate performance of both unconditional and conditional generation in our experiments.
The first objective is to maximize the likelihood of the current sentence given the previous sentence. Hence, we maximize the following: %cross-entropy loss that maximizes the probability of the input sequence $x$ compared to predicted sequence $\hat{x}=\texttt{softmax}(h_t)$:
\begin{align}
\mathbb{L}_{s2s} =  \sum_j \log P(w_{ij}| w_{i,<j}, s_{i-1})
\end{align}
where $s_i$=$\{w_{i1}, w_{i2},\ldots,w_{iT_i}\}$, and $T_i$ is the number of tokens of $s_i$.
%Generating the next word without any constraint is like choosing a random point in an infinite space of vocabulary choices.
%We reduce the complexity of search space for language modeling by jointly predicting RST relations.

To better guide the model with discourse context, we use the shared representations to predict RST relations at the same time.
% \hh{what specific thing did you do?}\dk{I assume that this is an effect of joint training. The sentence representations in language modeling are also learned by predicting the next relation.}
For each paragraph, we run the pre-trained RST parser~\cite{ji2014representation} and flatten the parse tree to obtain RST relations for each sentence $Y_i$=$(y_1,\ldots,y_{K_{i}})$, where $K_{i}$ is the number of discourse relations in $s_{i}$.
% We make a sequence of labels with the same number of word tokens by placing $y$ at the first word of EDUs and filling up the rest with a special token $o$: $Y'_i = (o,\ldots, o, \mathbf{y_1}, o,\ldots,\mathbf{y_{K_{i}}},o,\ldots, o)$.
We then make a label sequence over tokens in the sentence with by placing $y$ at the first word of EDUs and filling up the rest with a \textit{null} relation $o$: $Y'_i = (o,\ldots, o, \mathbf{y_1}, o,\ldots,\mathbf{y_{K_{i}}},o,\ldots, o)$.
% \tj{I think this looks weird since last discourse relations should be in between two last EDUs. So in general, the very last tokens in $Y'_{i}$ should be O}
We incorporate a sequence labeling objective by employing conditional random field \cite{Lafferty2001ConditionalRF} to find the label sequence that maximizes the score function for each sentence $s_i$: $\mathbf{S} (s_i, Y'_i) = \sum_{j=1}^{T_i-1} W^T_{y'_{j},y'_{j+1}}h_j + b_{y'_{j},y'_{j+1}}$ where $h_j$, $W$ and $b$ are the hidden representation of $w_{ij}$, weight matrix, and the bias vector corresponding to the pair of labels $(y'_i, y'_{i+1})$, respectively. %\in Y' \times Y'$,
For training, we maximize the conditional likelihood:
\begin{align}
\mathbb{L}_{CRF} &
% = \log (p(Y'_i|s_i))\\
% &= \mathbf{s} (Z,y) - \log (\sum_{y' \in \mathbb{Y}_x } e^{\mathbf{s} (Z,y')}) \\
& =  \mathbf{S} (s_i,y'_i) - \sum_{\mathbf{y} \in \mathbb{Y}_x} \log \mathbf{S} (s_i,\mathbf{y})
\end{align}
where $\mathbb{Y}_x$ represents all possible discourse label sequences. Decoding is done by greedily predicting the output sequence with maximum score. %by $y^* = \argmax_{y' \in \mathbb{Y}_x} \mathbf{s} (H,y)$
% \hh{This is not the greedy decoding. Needs fixed. You don't pick the best sequence out of all possible label sequence.}\dk{I don't understand your question.}
% \hh{I meant that this is a true objective but intractable. You approximate this by doing greedy decoding in practice, if I remember correctly.}
Both training and decoding can be computed using dynamic programming.
The final objective is represented as the sum of two objective functions:
%https://arxiv.org/pdf/1709.04109.pdf
\begin{align}\label{eq:loss}
\mathbb{L}_{\texttt{disc}} &= \mathbb{L}_{s2s} + \alpha * \mathbb{L}_{CRF}
\end{align}
where $\alpha$ is a scaling parameter to control the impact of CRF objective. The value is chosen empirically by searching based on validation set.
% To the best of our knowledge, this work is the first attempt of using LSTM-CRF architecture that incorporates discourse relations in language modeling.

% \dk{mention very sparse. change sentence to word level $h$}

%%%%%%%%%%%%%%%%%%%%
\subsection{Delta-driven \method}\label{sec:lmdelta}
%%%%%%%%%%%%%%%%%%%%
% Latent information is only based on the hierarchy of sentences.
% However, a sequence of sentences in the hierarchical models is limited to modeling topic transitions between sentences, while our aim is to characterize high-level flow which is independent of the topic.
In this model, we aim to utilize latent representations to characterize the flow between sentences. Specifically we define \textit{delta}, subtractions of hidden represenations of adjacent sentences as such latent information.
% Beyond hierarchical modeling of words and sentences, we assume that the model can also capture delta, a topic-invariant flow.
%The basic motivation of delta comes from the derivative $\frac{d}{dx}f(x)$ of a quadratic function $f(x)$, which is generally interpreted as an instantaneous rate of its change.
%Similarly, our delta function (i.e., subtraction) linearly captures the instant change between two adjacent sentence representations.
% \hh{don't understand this.}\dk{How about this?}\hh{hmmmmmm, we need to discuss.}
% -- as the third order (like a second-order derivative in an equation) information by a simple subtraction operation between two adjacent sentences.
Figure~\ref{fig:framework}(b) shows how we hierarchically model different levels of information: words, sentences, and \textit{deltas}.

% \hh{So if there is a previous work on three-layer hierarchical RNN, your model is just a subset of theirs?}\dk{Nope. delta is not another hierarchical representation but specifically operated to capture the flow.}\hh{If you have a RNN over sentences(same level as delta rnn), a RNN cell processes new sentence at a time and store information (or could be difference??) in the memory (that's what LSTM does). I guess one point is, given a sentence pair, couldn't you just apply a FFNN layer to get features, instead of subraction?}
Each word is encoded using a RNN encoder $g_{word}$. We take the last hidden representation of word as sentence embeddings $s_1,...,s_M$.
Similar to hierarchical RNN~\cite{lin2015hierarchical}, each sentence representation is encoded using another RNN encoder $g_{sent}$. %in $s_1$$\ldots$$s_M$
While discourse flow provides an explicit relation symbols, delta flow calculates a latent relation by subtracting previous representation $s_{i-1}$ from current representation $s_{i}$\footnote{Our experiment includes a comparison among other types of linear operations between sentences such as addition or a learnable function.}: %\hh{previous minus next? or next minus previous? see the equation below.}\dk{Thanks for pointing out this.}
\begin{align}
  d (s_{i-1},s_i) = d_{i-1} = s_{i} - s_{i-1}
\end{align}
Given a sequence of $M$-$1$ delta relations $d_1,...,d_{M-1}$ for a paragraph of $M$ sentences, we again encode them using another RNN encoder $g_{delta}$.
The model takes the word, sentence and delta information altogether to predict the next ($t$-th) word in the $m$-th sentence:
\begin{align}
h_t = f(h_{t-1}, x_t, s_{m-1},d_{m-2})
\end{align}
where $x_t$ is a word representation, $s_{m-1}$ is a sentence representation and $d_{m-2}$ is a delta information.
Note that sentence representation is from the previous sentence, and delta information is calculated by two previous sentences.
If there is no previous information given, the parameters are randomly initialized.
% For example, the delta information is calculated from the input sentences when decoding words of the third sentence.

%%%%%%%%%%%%%%%%%%%%%%%%%%%%%%%%%%%%%%%%%%%%%%%%%%%%%%
\section{Experiment}\label{sec:exp}
%%%%%%%%%%%%%%%%%%%%%%%%%%%%%%%%%%%%%%%%%%%%%%%%%%%%%%
% We test the usefulness of linguistic and latent relations in partial paragraph generation task.
% \tj{Ordering task on Table~\ref{tab:results} is out. You should modify your context.}
% We would like to answer four questions in our experiment: (1) Are linguistic and latent relations useful for producing coherent paragraph? (2) Is the \textsc{subtract} operation more efficient to capture coherence compared to other operations such as addition? (3) Are discourse or delta relations more effective as the number of paragraph lengths to predict increase?
% %or across different domains? 
% (4) How does human perform on the tasks compared to the models?
%We begin by describing the data collection process for the purpose of our new tasks.
Due to the absence of goal-oriented language generation task, we collect paragraph data and define a new task of generating partial text of a paragraph given some context.

%%%%%%%%%%%%%%%%%%%%
\subsection{Data}\label{sec:collection}
%%%%%%%%%%%%%%%%%%%%
\begin{table}[th!]
\small
\begin{center}
\caption{\label{tab:dataset} Number of paragraphs in our dataset. \vspace{-3mm}}
\centering\vspace{0mm}
\begin{tabular}{@{}l|rrr@{}}
\toprule
&  \textbf{Train}&  \textbf{Valid}&  \textbf{Test} \\
\midrule
% \texttt{News}&	3,862&	216&	216\\
\texttt{Papers}&	16,173&	899&	899\\
\texttt{SciFi}  &	157,031&	8,724&	8,724\\
\texttt{Fantasy}  &	317,654&	17,649&	17,649\\
% - \texttt{Romance}&	1,026,798&	62,371&	62,371\\
\bottomrule
\end{tabular}
\end{center}\vspace{-3mm}
\end{table}

% We describe our data used in the experiment.
% \hh{please don't refer news corpus as News before you define (even in the prelim analysis), if you define \texttt{News} for the first time here.}\dk{good point. deleteted all of previous ones now.}
% In \S\ref{sec:motivation}, we show that different domains or positions of text have their own meaning variations between sentences.
% For example, paragraphs in News articles likely deliver a fact with less space so repeating similar phrases at first and last, while paragraphs in a novel book are more narrative and sequential.
We collect paragraphs from three different domains: \texttt{Papers} are paragraphs extracted from academic manuscripts in computer science domain from the PeerRead \cite{kang2018peerread}, and \texttt{Fantasy} and \texttt{SciFi} are paragraphs of two frequent categories extracted from the BookCorpus  \cite{moviebook}, where paragraphs are extracted using the line breaker in the dataset.

% \begin{itemize}[noitemsep,topsep=0pt,leftmargin=*]
% % \item \texttt{News}: are paragraphs from the New York Times Annotated Corpus~\cite{sandhaus2008new}. 
% % We parse body text using $<$p$>$ tag in the XML files as a delimiter of sentences. %\hh{paragraph, right?}\dk{Nope. They don't have paragraph delimiter so I just assume every sentences as on paragraph.}
% % and choose the text as a paragraph if they have multiple sentences.
% \item  %\hh{Why only three genres?}\dk{The dataset only has three categories lol.}.
% % The paragraphs are extracted using the science-parser ~\footnote{\url{github.com/allenai/science-parse}}.
% % The dataset also includes the parsed PDFs with  so we extract paragraphs using line-break delimiters.
% %\hh{Is this pdf parsing done by your peerread dataset, or you did it for this particular task?}\dk{Paragraph extraction is done by me. Corrected.}
% \item . 
% \end{itemize}

We only use paragraphs whose lengths are from 4 to 7, in order to measure the performance change according to paragraph length.
The dataset is randomly split by 0.9/0.05/0.05 for train, valid, and test set, respectively. 
Table~\ref{tab:dataset} shows the numbers of paragraphs for each domain. 
All paragraphs are parsed into RST trees using the state-of-the-art discourse parser by~\citet{ji2014representation}.
%\hh{Useless information. nOt needed.}\dk{Agreed and deleted it.}
% \footnote{Our parsing took more than three weeks}
% We summarize other heuristics to filter out noisy paragraphs in the Appendix.

%%%%%%%%%%%%%%%%%%%%
\subsection{\textit{Bridging}: Partial Paragraph Generation}\label{sec:tasks}
%%%%%%%%%%%%%%%%%%%%
% Text generation without any context is like finding your friend in an infinite space. 
% Instead, we reduce the search space of text generation by providing partial information of a long text similarly as \citet{kiddon2016globally} provides a handful keywords for recipe generation.  
%\hh{what? what do readers learn from this sentence?}
%\hh{I don't understand what you mean by reduce space. Conditioned generation is still decoding over infinite space.}
%\dk{Think about general language modeling. There is no other way to evaluate it except perplexity, because LM is basically generation from scratch. My intention here is to suggest more goal-oriented language modeling by providing partial information and target.}
%\dk{I rewrote it below. How about now?}
%Language models are often evaluated by the predictability of words using perplexity. 
%However, a long text such as a paragraph is difficult to evaluate its coherence only using the perplexity. 
%Instead,
We evaluate our models on partial text generation task; given a partial information (e.g., some sentences), producing the rest of text.
% Similarly, \cite{kiddon2016globally} provides a handful keywords for recipe generation.

\begin{figure}[ht!]
\vspace{0mm}
% \centering
% \small
\caption{\label{fig:tasks} Bridging task: given [1] and [4] sentences, guessing [2,3] sentences (red, underlined). \vspace{-2mm}}\vspace{0mm}
\begin{tabularx}{\linewidth}{@{}X@{}}
\hline
[1] Inside the club we moved straight for the bar.
{\color{red}
\ul{[2] Devlin ordered a beer for himself and a glass of my favorite wine for me.}
%{\color{vlgray}Devlin ordered a} {\color{black}beer} {\color{vlgray}for himself and a glass of my favorite} {\color{black}wine} {\color{vlgray}for me.}
% {\color{vlgray}I} {\color{black}love} {\color{vlgray}that I didn't have to} {\color{black}tell} {\color{vlgray}him what I wanted.}
\ul{[3] I love that I didn't have to tell him what I wanted.}
}
[4] He knew me well and always thought about what I wanted or needed, in and out of bed.\\
% \bottomrule
\hline
\end{tabularx}
%\textit{\textbf{Seeds}}: [{\color{blue}beer}, {\color{blue}wine}], [{\color{red}love}, {\color{red}tell}]\\
% \newline
% \vspace*{1mm}
% \newline
% \begin{tabular}{@{}>{\arraybackslash}p{\linewidth}@{}}
% %\toprule
% \textit{\textbf{Ordering}}\\
% \midrule
% Inside the club we moved straight for the bar.
% \underline{ - {\color{red}(3)} - {\color{red}(2)} -}
% He knew me well and always thought about what I wanted or needed, in and out of bed.\\
% \midrule
% (2) I love that I didn't have to tell ...\\
% (3) Devlin ordered a beer for himself and a ...\\
% \midrule
% %\bottomrule
% \end{tabular}\vspace{0mm}
% \vspace{-2mm}
\end{figure}

%\hh{Better make a subsubsection for each task. Also divide the following paragraph accordingly in two parts.}\dk{seems a good suggestion. I made two paragraphs instead, because I don't like subsubsection}
% We introduce two tasks to evaluate our models : \textit{Bridging} and \textit{Ordering}. %using the paragraph dataset we collected

% \hh{How about below?}
%\paragraph{Bridging} needs to generate hidden intermediate sentences by connecting or bridging the first and last sentence in a consistent way (See Figure~\ref{fig:tasks} (top)).
Figure~\ref{fig:tasks} shows our bridging task.
It requires a generation of masked sentences in the middle of a paragraph given the first and the last sentences. %of the paragraph. 
% It requires connecting or bridging information of a paragraph in a consistent way.
If only the first sentence is given, the generation can be too divergent.
The existence of the last sentence makes the generation more coherent and converged to some point. 
%we need to produce appropriate text that coherently bridges the first and last sentences.

We evaluate it with one hard and one soft automatic metrics: METEOR (M)~\cite{banerjee2005meteor} and VectorExtrema (VE) \cite{liu2016not} by calculating cosine similarity of averaged word embeddings \cite{pennington2014glove}, and human performance.
\subsection{Models and Setup}\label{sec:models}
%%%%%%%%%%%%%%%%%%%%

We compare various baseline seq2seq models which encode the context; a concatenated first and last sentences, and decode the intermediate words: 
\textsc{\textbf{S2S}} is attentional seq2seq model \cite{bahdanau2014neural}, and
\textsc{\textbf{HS2S}}: is a hierarchical version of the S2S by combining two baselines: HRNN~\cite{lin2015hierarchical} hierarchically models sequence of words and sentences, and HRED~\cite{serban2017hierarchical,sordoni2015hierarchical} encodes the given context and decodes the words.
\textbf{\method} \textbf{(delta/disc.)} is our proposed language model with delta and discourse relations, respectively. 

We find the best hyper-parameters on validation set using grid search.
Here are the final parameters used:
$32$ for batch size, $25$ for maximum sentence length, $300$ for word embedding size initialized by GloVe~\cite{pennington2014glove}, $1$ LSTM layer \cite{hochreiter1997long} with $512$ size, clipping by $0.25$, $0.2$ learning rate and $0.5$ decay rate with Adagrad~\cite{duchi2011adaptive} optimizer, and $50,000$ for the vocabulary size.
The total number of distinct discourse relations is $44$.

%%%%%%%%%%%%%%%%%%%%%%%%%%%%%%
\subsection{Results}
%%%%%%%%%%%%%%%%%%%%%%%%%%%%%%

\begin{table}[ht!]
\vspace{0mm}
\small
\begin{center}
\caption{\label{tab:results} Performance on bridging task. %(top) and ordering (bottom) 
\textbf{M}ETEOR and \textbf{V}ector\textbf{E}xtrema are used.
The higher the better.\vspace{-3mm}}
\centering
\begin{tabular}{@{}r !{\vrule width0.8pt} cc cc cc @{}}
% !{\vrule width0.8pt} @{}x{0.6cm}@{}x{0.6cm}@{}x{0.6cm}@{}x{0.7cm}@{}
\toprule
&       \multicolumn{2}{c}{\texttt{Papers}} &   \multicolumn{2}{c}{\texttt{SciFi}}&   \multicolumn{2}{c}{\texttt{Fantasy}}\\
%&   \multicolumn{3}{c}{\texttt{Romance}}
\midrule
&  \textbf{M} &   \textbf{VE} & \textbf{M} &    \textbf{VE} & \textbf{M}  &   \textbf{VE}\\
% & \textbf{B} & \textbf{M} & \textbf{R} &   \textbf{VE} 
\midrule
\textsc{\textbf{S2S}}  &  3.7 & 56.3 & 3.5 & 71.0 & 3.3  & 66.3 \\
\textsc{\textbf{HS2S}}  &  3.7 &  54.7 & 3.4 &  \textbf{73.0} & 3.0 & 69.7\\
% \textbf{H-RED} \\
\midrule
\textbf{\method} (delta)  &  3.1 &  \textbf{58.5} &  3.6 & 69.7 &3.6 & \textbf{73.9} 
\\
\textbf{\method} (disc.) &   \textbf{4.0} &  57.2 & \textbf{4.2} &  70.3 & \textbf{3.9} &  71.8 
\\
\bottomrule
\end{tabular}
\end{center}
% \begin{center}
% \begin{tabular}{@{}r !{\vrule width0.8pt} @{}x{0.8cm}@{}x{0.8cm}@{}x{0.8cm}@{} !{\vrule width0.8pt} @{}x{0.8cm}@{}x{0.8cm}@{}x{0.8cm}@{} !{\vrule width0.8pt} @{}x{0.8cm}@{}x{0.8cm}@{}x{0.8cm}@{} !{\vrule width0.8pt} @{}x{0.8cm}@{}x{0.8cm}@{}x{0.8cm}@{}@{}}
% \toprule
% &   \multicolumn{3}{c}{\texttt{News}} &   \multicolumn{3}{c}{\texttt{Papers}} &   \multicolumn{3}{c}{\texttt{SciFi}}&   \multicolumn{3}{c}{\texttt{Fantasy}}\\
% \midrule
% %&   \multicolumn{3}{c}{\texttt{Romance}}
% & \textbf{N} & \textbf{P@k} & \textbf{A(P)} & \textbf{N} & \textbf{P@k} & \textbf{A(P)}& \textbf{N} & \textbf{P@k} & \textbf{A(P)}& \textbf{N} & \textbf{P@k} & \textbf{A(P)}\\
% %& \textbf{N} & \textbf{P@k} & \textbf{A(P)}
% \midrule
% \textsc{\textbf{S2S}} & \textbf{0.92} & \textbf{0.91} & \textbf{0.77}& 0.84 & 0.82 & 0.62& 0.89 & 0.87 & \textbf{0.66} & 0.90 & 0.88 & 0.67\\
% \textsc{\textbf{HS2S}} & 0.90 & 0.89 & 0.75& \textbf{0.86} & \textbf{0.84} & \textbf{0.62}& 0.88 & 0.87 & 0.65 & 0.90 & 0.89 & 0.67\\
% \midrule
% \textsc{\textbf{\method(\implicit)}} &  0.90 & 0.88 & 0.75& 0.85 & 0.83 & 0.61& \textbf{0.89} & \textbf{0.87} & 0.65 & \textbf{0.90} & \textbf{0.89} & \textbf{0.67}\\
% \textsc{\textbf{\method(\explicit)}} &0.91 & 0.90 & 0.76 & 0.85 & 0.82 & 0.61& 0.87 & 0.85 & 0.64 & 0.89 & 0.87 & 0.65\\
% \bottomrule
% \end{tabular}
% \end{center}\vspace{0mm}\vspace{0mm}
\vspace{0mm}
\end{table}

%\hh{what do you mean here? you didn't think of modeling together from the beginning, because they are different.}\dk{I think this is now clear and fixed.}
% For ordering, we encode the context, calculate likelihoods of given intermediate sentences and then choose the most likelihood sentence for every permutation of middle sentences' orders.

In Table \ref{tab:results}, both discourse and delta driven \method outperform the baseline models across most of the metrics except for VecterExtrema on \texttt{SciFi}. 
% \paragraph{More benefits are obtained from \method as data size increases.}
Especially, as the number of training size increases (\texttt{Papers}$<<$\texttt{SciFi}$<$\texttt{Fantasy}), the improvements gained from the \method become bigger. 
This is probably because the model learns more information of the (discourse or latent) relations from the larger data.
% Among different domains, \method(discourse) achieves better performance on \texttt{SciFi} and \texttt{Papers}, while \method(delta) on \texttt{News}.

%For ordering, \method(\implicit) outperforms the baselines and \method(\explicit) on the larger size of datasets (e.g., \texttt{SciFi}, \texttt{Fantasy}), while both \method models don't help much in less data setting (e.g., \texttt{News}, \texttt{Papers}).

\begin{figure}[h!]
\vspace{-2mm}
\small
\begin{floatrow}
\capbtabbox[3.2cm][]{%
\begin{tabular}{@{}r@{\hskip 1mm}|@{\hskip 1mm}c@{\hskip 3mm}c@{}}
\toprule
 &M&	VE\\
\midrule
\textsc{subtract} & 3.35 & \textbf{67.20}\\
\textsc{add} &  \textbf{3.45} & 65.35 \\
\textsc{mlp} & 3.32 & 62.97\\
\bottomrule
\end{tabular}\vspace{0mm}
}{%
  \caption{\label{tab:delta_comparison} Comparison of different delta functions. }%
}
\hspace{-3mm}
\ffigbox[3.7cm][]{%
\includegraphics[trim={2mm 65mm 4mm 2mm},clip,width=.95\linewidth]{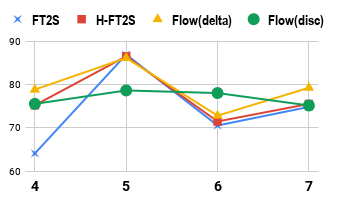}
\includegraphics[trim={4mm 2mm 2mm 2mm},clip,width=.95\linewidth]{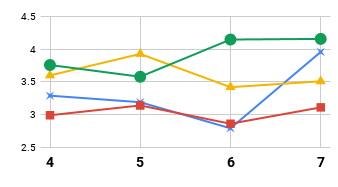}
\vspace{-2mm}
% \subfloat[][\small{VectorExtreme}]{}
}{%
  \caption{Comparison of paragraph lengths. Best viewed in color.}\label{fig:length}%
}
\end{floatrow}
\vspace{-3mm}
\end{figure}

\begin{table*}[ht!]
\vspace{0mm}
\centering
\small
\caption{\label{tab:generated}
An example paragraph and predicted texts in \texttt{Fantasy} dataset.
Given \textsc{\textbf{First}} and \textsc{\textbf{Last}} sentences, the models generate middle sentences (e.g., {{\color{blue}{\tiny\textbf{[M1]}}}} $\rightarrow$ {{\color{blue}{\tiny\textbf{[M2]}}}}..). \textsc{\textbf{Ref}} and \textsc{\textbf{Human}} are reference middle sentences and sentences written by human annotator, respectively. 
Please find more examples in the appendix.
}
\vspace{0mm}
\begin{tabularx}{\linewidth}{@{}X@{}}
\toprule
% Source: \texttt{Fantasy} with length $6$ paragraphs at $432$-th testing sample\\
% \midrule
\textsc{\textbf{First}}: Satyrs never wear armor, including helmets, Newel began, using his hands expressively.\\
\textsc{\textbf{Last}}: Anyhow, as we actors were laying siege, a big chunk of the battlement dislodged from atop the tower.\\
\midrule
\textsc{\textbf{Ref}}: {{\color{blue}{\tiny\textbf{[M1]}}}} "But years ago I was in a play, and the helm was part of my costume. {{\color{blue}{\tiny\textbf{[M2]}}}} During the big battle scene, a few of us were assailing a castle. {{\color{blue}{\tiny\textbf{[M3]}}}} We had quite a set. {{\color{blue}{\tiny\textbf{[M4]}}}} The main tower must have been fifteen feet tall, fashioned from real stone.\\
\textsc{\textbf{Human}}: {{\color{blue}{\tiny\textbf{[M1]}}}} Actually he needed to wear any protectors to prevent him from a big accident. {{\color{blue}{\tiny\textbf{[M2]}}}} We planned to make a prank cam to make him wear those always. {{\color{blue}{\tiny\textbf{[M3]}}}} "I have a good idea," Newel kept talking continuously. {{\color{blue}{\tiny\textbf{[M4]}}}} "Let's play a role like we are under the attack.\\
\midrule
\textsc{\textbf{S2S}}: {{\color{blue}{\tiny\textbf{[M1]}}}} he's a good man {{\color{blue}{\tiny\textbf{[M2]}}}} the UNK, the one who's a man who's a man and the other {{\color{blue}{\tiny\textbf{[M3]}}}} and the other, the one who 's a good friend {{\color{blue}{\tiny\textbf{[M4]}}}} he's a good man\\
\textsc{\textbf{HS2S}}: {{\color{blue}{\tiny\textbf{[M1]}}}} i'm not sure that," he said {{\color{blue}{\tiny\textbf{[M2]}}}} i'm not sure that i'm not sure {{\color{blue}{\tiny\textbf{[M3]}}}} i'm not sure that i'm not a fool {{\color{blue}{\tiny\textbf{[M4]}}}} "i'm not sure that," he said\\
\textsc{\textbf{\method(delta)}}: {{\color{blue}{\tiny\textbf{[M1]}}}} he's a good man {{\color{blue}{\tiny\textbf{[M2]}}}} i'm not sure what to do {{\color{blue}{\tiny\textbf{[M3]}}}} i'm not sure that i'm not going to be a vampire {{\color{blue}{\tiny\textbf{[M4]}}}} he's a good man\\
\textsc{\textbf{\method(disc.)}}: {{\color{blue}{\tiny\textbf{[M1]}}}} perhaps they were not quite good, but he was not a master, and they were the most powerful {{\color{blue}{\tiny\textbf{[M2]}}}} the only way to do not like a little, but i' d been in the world {{\color{blue}{\tiny\textbf{[M3]}}}} "you're right," he said "i am not a fool you're here {{\color{blue}{\tiny\textbf{[M4]}}}} you're going to be a bit more than the other\\
\bottomrule
\end{tabularx}\vspace{0mm}
\end{table*}

% \tj{You also need to think about the result from \ref{tab:delta_comparison}. It doesn't guarantee that the subtraction is the best option for the opreation...}
Table~\ref{tab:delta_comparison} shows performance comparison among different delta operations: \textsc{subtract}, \textsc{add}, and \textsc{mlp} which is a multi-layer perceptron network.
All scores are macro-averaged across datasets.
While \textsc{add} shows good performance on METEOR, \textsc{subtract} does on the soft metric (i.e., VecExt), indicating that subtraction can help the model capture the better semantics than the other functions. 
%the topic-invariant flow. 
%Especially in \texttt{Fantasy} dataset, \textsc{subtract} achieves +4.0 and +7.1 VE scores against \textsc{add} and \textsc{mlp}, respectively. 
% \tj{I didn't get rid of the ordering result from \ref{fig:length}. You may want to get rid of it...}
Figure~\ref{fig:length} shows how performance changes on \texttt{Fantasy} as the paragraph lengths increase. 
Both of \method achieve more improvements when generating longer paragraphs. 
Especially, discourse relations achieve the best performance at length 6 and 7.
% For ordering, \method (delta) shows consistent improvements over the baselines, and the gap increases as the length increases. 
% \tj{I don't think the last statement is true. The biggest gap from (b) is length = 4, which is the smallest length.}

\begin{figure}[h!]
\vspace{0mm}
\centering
{
% \includegraphics[trim={2mm 1mm 9mm 2mm},clip,width=.5225\linewidth]{figs/human_order_length.png}
% \subfloat[][\small{Ordering by length}]{}
% \includegraphics[trim={9mm 40mm 9mm 2mm},clip,width=.5\linewidth]{figs/human_order_dataset.png}
% \subfloat[][\small{Ordering by dataset}]{}\\\vspace{0mm}
\subfloat[][By paragraph lengths]{\includegraphics[trim={2mm 2mm 9mm 7mm},clip,width=.47\linewidth]{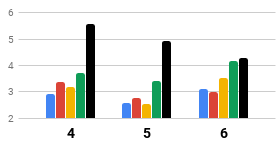}}
\quad
\subfloat[][By domains]{\includegraphics[trim={2mm 2mm 9mm 7mm},clip,width=.47\linewidth]{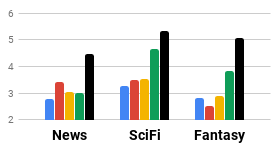}}
}\vspace{0mm}
\caption{Comparison (METEOR) with human performance (black bars): S2S (blue), HS2S (red), Flow:delta (yellow), and Flow:disc. (green). Best viewed in color.\vspace{0mm}}\label{fig:human}
\end{figure}

We conduct a comparison with human performance (See Figure~\ref{fig:human}).
We randomly choose 100 samples per dataset and per paragraph length and ask an annotator to perform the bridging task on the final 1,000 samples.  %\footnote{We exclude \texttt{Papers} dataset and length 7 paragraphs because they are too difficult even for human annotators.}. 
% An annotator performs bridging tasks on the final 1,000 samples.
% Figure~\ref{fig:human} shows a performance comparison of our models against human annotators.
% Surprisingly, most of models achieve better-than-human performance in ordering task except length 4 paragraphs.
Human outperforms the models by large margins. 
\method with discourse relations outperforms the \method with latent relations and other baselines by a large margin. 
As the paragraph length increases or more data is trained, discourse relations become more useful. 
% We include example paragraphs produced by different models in Appendix.

Table~\ref{tab:generated} shows an example paragraph with text produced by the models as well as reference and human annotation. 
Given only the partial context (i.e., first and last sentences), bridging task is very challenging even for human. 
The reference sentences and human annotations are semantically very different indeed.
% Especially, when the context is very less, the model predictions are dominated by frequent responses in the training corpora. % such as the first case
Among the latent models, \method (delta) produces more coherent flow of text compared to S2S and HS2S.
Surprisingly, \method (discourse) enables generating more diverse sentences with a bit of coherence, because each sentence is generated based on the representation conditioned on the predicted RST discourse relation.

% \subsection{Future Directions.}

% \begin{figure}[htbp!]
% \centering
% \subfloat[][\small{NDCG}]{\includegraphics[draft,trim={0mm 0mm 0mm 0mm},clip,width=.3\linewidth]{example-image-c}}
% \caption{Comparison between discourse and delta relations in continuous space. We use PCA projection.}\label{fig:comparison_pca}
% \end{figure}\vspace{0mm}
% Figure~\ref{fig:comparison_pca} shows comparison between discourse and delta relations. 

% \dk{TODO: Human eval}

%%%%%%%%%%%%%%%%%%%%%%%%%%%%%%%%%%%%%%%%%%%%%%%%%%%%%%
\section{Conclusion and Discussion}
%%%%%%%%%%%%%%%%%%%%%%%%%%%%%%%%%%%%%%%%%%%%%%%%%%%%%%
We explore two forms of inter-sentential relations: linguistic relation such as discourse relations and a latent representation learned from the text.
The proposed models for both relations achieve significant improvements over the baselines on partial paragraph generation task.
% We believe that our works helps understand the relation-driven control of text planning for generation task.

Despite the empirical effectiveness and difference between the linguistic and latent relations, they are not directly aligned for comparison.
A potential direction for future study is to directly couple them together and see whether one form contains the other, or vice versa.
Another direction is to check their effectiveness on top of the recent pre-trained language models.

\section*{Acknowledgements}
We also thank Jason Weston, Dan Jurafsky, and anonymous reviewers for their helpful comments.

\bibliography{paragen}
\bibliographystyle{acl_natbib}

\renewcommand*\appendixpagename{\Large Appendices}

\clearpage

\begin{appendix}
\label{sec:appendix}

% \vspace{150mm}

\section{Details on data processing}
For each dataset, we preprocess each paragraph as follows:
\begin{itemize}[noitemsep,topsep=0pt,leftmargin=*]
\item The sentences which length is less than 5 or higher than 25 are filtered out to remove too short or too long sentences. 
\item Due to too much noise from News and papers corpus, we make bit aggressive filters. The paragraphs whose last sentence ends with all capital words are removed to filter out the articles with reporter's name or other meta information (e.g., location, press name). Also, paragraphs whose last sentences don't end with sentence-ending marks (e.g., “.”, “!”, “?”) are also filtered out. 
\item If any adjacent sentences in a paragraph is identical, we exclude the paragraph. All the duplicate paragraphs are also removed. 
\item We ignore the paragraph that fails to be parsed by our discourse (i.e. RST) parser. The detail of the parsing would be described in the next section. During the RST parsing, some Stanford dependency parses contain UNK token  that is mismatched with our tokenizer (i.e. nltk's word tokenizer). Then, we also ignore such cases (only 1.5\% of entire dataset).
\end{itemize}
% \section{More Examples}
% Table~\ref{tab:generated_appendix} and \ref{tab:generated} shows another example produced by our models and human annotator.

\section{Theoretical difference between linguistic and latent relations}
% Before empirical comparison of the two forms of relations, w
We briefly summarize the fundamental differences of the two relation forms:
\begin{itemize}[noitemsep,topsep=0pt,leftmargin=*]
\item While labels in linguistic relations are interpretable, accuracy of the labels highly depends on the performance of the discourse parser. On the other hand, latent representations with delta operation do not suffer from out-of-domain or accuracy problems that external parsers may bring in. 
\item Linguistic relations can hold over long distances with many things in between (e.g., Solutionhood), while the latent ones are always immediately adjacent.  
\item Linguistic ones are fairly coarse-grained and non-continuous, often making them inapplicable to other continuous models (e.g., a neural network), while latent ones are by definition continuous, always making them applicable.
\item Linguistic relations are often ambiguous or unclear, while latent ones can easily hybridize and represent two more relations at the same time, at the cost of being indefinable.  
\end{itemize}

\noindent
\begin{minipage}{0.5\textwidth}
% \captionof{figure}{Big Figure}
% \begin{table*}[ht!]
\vspace{0mm}
% \small
\captionof{table}{\label{tab:generated_appendix}Example texts produced by different models. Given \textsc{\textbf{First}} and \textsc{\textbf{Last}} sentences, the models generate middle sentences (e.g., {{\color{blue}{\tiny\textbf{[M1]}}}} $\rightarrow$ {{\color{blue}{\tiny\textbf{[M2]}}}}$\rightarrow$ {{\color{blue}{\tiny\textbf{[M3]}}}}..). \textsc{\textbf{Ref}} and \textsc{\textbf{Human}} are reference middle sentences and sentences written by human annotator, respectively. 
}
\centering
\begin{tabularx}{\textwidth}{@{}X@{}}
\toprule
% Source: \texttt{SciFi} with length $5$ paragraphs at $1086$-th testing sample\\
% \midrule
\textsc{\textbf{First}}: Okay, 'Molinari sighed.\\
\textsc{\textbf{Last}}: His voice rose to a shout of anger.\\
\midrule
\textsc{\textbf{Ref}}: {{\color{blue}{\tiny\textbf{[M1]}}}} I'll get up; just leave me alone, will you, for chrissake?' {{\color{blue}{\tiny\textbf{[M2]}}}} He stirred about, struggling to get from the bed. {{\color{blue}{\tiny\textbf{[M3]}}}} Okay - I'll get up; will that satisfy you?'\\
\textsc{\textbf{Human}}: {{\color{blue}{\tiny\textbf{[M1]}}}} What? Is this everything you want me to talk to? he couldn't stop laughing this situation. {{\color{blue}{\tiny\textbf{[M2]}}}} I don't want to talk with you anymore, Molinari said. {{\color{blue}{\tiny\textbf{[M3]}}}} You make me crazy. You disappoint me so badly!!\\
\midrule
\textsc{\textbf{FT2SEQ}}: {{\color{blue}{\tiny\textbf{[M1]}}}} you are a few times and you are not a good person {{\color{blue}{\tiny\textbf{[M2]}}}} she has a good idea and the others are not a bit of pleasure {{\color{blue}{\tiny\textbf{[M3]}}}} he was a little bit of his own life\\
\textsc{\textbf{H-FT2SEQ}}: {{\color{blue}{\tiny\textbf{[M1]}}}} the two of the men are not a UNK {{\color{blue}{\tiny\textbf{[M2]}}}} the two of them are not to be able to make a small map of the ship {{\color{blue}{\tiny\textbf{[M3]}}}} he had to be a man who had been a man who had been a man who had been a man\\
\textsc{\textbf{\method(delta)}}: {{\color{blue}{\tiny\textbf{[M1]}}}} he's a very large man, and a man with a very long way {{\color{blue}{\tiny\textbf{[M2]}}}} i'm not sure that i'm not going to be able to get rid of it {{\color{blue}{\tiny\textbf{[M3]}}}} he had to be a child and a man\\
\textsc{\textbf{\method(disc.)}}: {{\color{blue}{\tiny\textbf{[M1]}}}} you're going to be a good time {{\color{blue}{\tiny\textbf{[M2]}}}} it's a lot of the people, and the other worlds is a simple place to be in the universe that {{\color{blue}{\tiny\textbf{[M3]}}}} you're not to be a friend\\
\bottomrule
\end{tabularx}
\end{minipage}

\end{appendix}

\end{document}